\documentclass{article}

\usepackage{amsmath,amssymb}
\usepackage{times}
\usepackage{graphicx}
\usepackage{subfigure}
\usepackage{natbib}
\usepackage{algorithm}
\usepackage{algorithmic}
\usepackage{lipsum}
\usepackage{todonotes}
\usepackage{booktabs}
\usepackage{hyperref}

\usepackage[final]{nips_2018}

\author{
    Samuel G. Finlayson\thanks{These authors contributed equally.} \\
  Department of Systems Biology\\
  Harvard Medical School\\
  Boston, MA 02130 \\
  \texttt{samuel\_finlayson@hms.harvard.edu} \\
   \And
  Hyunkwang Lee\textsuperscript{*} \\
  School of Engineering and Applied Sciences \\
  Harvard University \\
  \texttt{hyunkwanglee@seas.harvard.edu} \\
   \And
  Isaac S. Kohane \\
  Department of Biomedical Informatics\\
  Harvard Medical School\\
  Boston, MA 02130 \\
  \texttt{isaac\_kohane@hms.harvard.edu} \\
  \And
  Luke Oakden-Rayner\\
  School of Public Health \\
  The University of Adelaide \\
  Adelaide, SA 5000 \\
  \texttt{luke.oakden-rayner@adelaide.edu.au}
}

\title{Towards generative adversarial networks as a new paradigm for radiology education}

\begin{document}

\maketitle

\begin{abstract}

Medical students and radiology trainees typically view thousands of images in order to "train their eye" to detect the subtle visual patterns necessary for diagnosis. Nevertheless, infrastructural and legal constraints often make it difficult to access and quickly query an abundance of images with a user-specified feature set. In this paper, we use a conditional generative adversarial network (GAN) to synthesize $1024\times1024$ pixel pelvic radiographs that can be queried with conditioning on fracture status. We demonstrate that the conditional GAN learns features that distinguish fractures from non-fractures by training a convolutional neural network exclusively on images sampled from the GAN and achieving an AUC of $>0.95$ on a held-out set of real images. We conduct additional analysis of the images sampled from the GAN and describe ongoing work to validate educational efficacy.

\end{abstract}

\section{Introduction}

Medical trainees must view thousands of images over the course of their training in order to achieve competency in visual diagnosis \citep{wang2012competencies}. Nevertheless, medical images pose unique legal and technical challenges that make it difficult to widely disperse images for training between sites or rapidly and simply query data for training purposes within a site \citep{ching2018opportunities}.

In this paper, we propose medical generative adversarial networks (GANs) as a method for developing (human) training tools in the field of radiology. Utilizing the architecture developed by \citet{karras2017progressive}, we train a conditional GAN over fractured and non-fractured hip radiographs using the dataset utilized in \cite{gale2017detecting}. We sample conditional images at 1K resolution, and perform in silico experiments to validate their capacity to re-capitulate key visual features distinguishing fractures from healthy hips. Finally, we describe human experiments that are under development at time of workshop submission, but will more fully elucidate the comparative benefit of GAN-generated images in training medical professionals.

\section{Background}

\subsection{Generative Adversarial Networks}

Generative Adversarial Networks (GANs) are a framework proposed by \citet{goodfellow2014generative} to estimate generative models over rich data distributions. The key insight of the GAN framework is to pit a generator model $G$ against a discriminator model $D$: The discriminator is optimized to distinguish real images from synthetic images, whereas the generator is optimized to reproduce the true data distribution $p_{data}$ by generating images that are difficult for the discriminator to differentiate.

Taken together, $D$ and $G$ can be said to engage in a two-player min-max game with the following objective function 
\begin{equation}\label{eq:GAN_ori}
\begin{aligned}
\min_{G} \max_{D} V(D,G) = \; & \mathbb{E}_{x \sim {p_{data}}} [\log D(x)] \; + \\
& \mathbb{E}_{z \sim {p_{z}}} [\log(1 - D(G(z)))],
\end{aligned}
\end{equation}
where $x$ is a real image from the true data distribution $p_{data}$, and $z$ is a noise vector sampled from distribution $p_{z}$ (\emph{e.g}., uniform or Gaussian distribution). 

Conditional GANs ~\cite{gauthier2014conditional, mirza2014conditional} are an extension of the GAN game, where both the generator and discriminator receive additional conditioning variables $c$, yielding $G(z,c)$ and $D(x,c)$. This formulation allows $G$ to generate images conditioned on variables $c$. Conditional GANs have been used to extend the GAN framework to many applications, including generating images based on specific features such as text descriptions. \citet{karras2017progressive} proposed a GAN architecture that starts with small images and progressively adds layers to grow them to higher resolutions throughout training.

There has been considerable debate in recent years pertaining to the extent to which GANs are truly "distribution learners" and their utility for tasks such as data augmentation \citep{santurkar2017classification, li2018limitations, antoniou2017data}. This may imply that in the immediate term, some degree of caution should be taken when using GANs in clinical applications.  

\subsection{Motivation and Related Work}

Medical imaging poses a paradoxical pair of challenges from the perspective of information technology: On the one hand, medical images create a \textit{big data problem}: millions of medical images are ordered each year within the United States alone, each of which must be processed, interpreted by medical experts, and stored. Nevertheless, most who work with medical data suffer from a \textit{small data problem}: medical images are generally siloed and incompletely annotated, making it difficult in many cases for individual researchers and medical trainees to compile sufficient examples for human or machine learning tasks \citep{ching2018opportunities}.

A number of attempts have been made to apply GANs to medical imaging datasets \citep{nie2017medical, dar2018image, wolterink2017deep, hu2017freehand, guibas2017synthetic, korkinof2018high, beers2018high}, including more recently the use of conditional GANs for data augmentation in training medical machine learning classifiers \citep{frid2018synthetic, wu2018conditional}. Seemingly unexplored, however, is the possibility of utilizing GANs as a source of training images for humans.

There are several technical and practical reasons for which clinical education may be an ideal deployment setting for medical GANs. Educational environments are comparatively lower stakes than clinical settings, and do not need connect with EHR or PACS systems. As the data backbone for more advanced educational tools, GANS could provide an exceptionally large set of human training examples with comparatively small and entirely self-contained storage requirements. To boot, GANs could in theory allow for human-driven or programmatic conditioning on clinical attributes, which could be used to build rapid and interactive learning experiences.

Educational tools need not cover the full breadth of a clinical domain to provide utility to a learner; for example, consider the frequently visited webpage "Normal Chest X-Ray: Train Your Eye", which consists entirely of 500 examples of \textit{normal} chest-xrays (see: \url{http://www.chestx-ray.com/index.php/education/normal-cxr-module-train-your-eye}). This could mean that medical GANs could be useful in an educational setting even if they only learn to sample from a small subset of the data distribution.

\section{Methods}

\subsection{Data and Model}

We trained a generative adversarial network to generate high resolution ($1024 \times 1024$) hip images conditioned on fracture status.

Institutional approval was granted for all stages of this project. Data for this study consisted of 11,734 pelvic X-ray images of which 3,289 images contain fractures. Data were obtained from the clinical radiology archive at The Royal Adelaide Hospital as described in \citet{gale2017detecting}. In brief, data were preprocessed using a series of convolutional neural networks to identify and extract the region of the pelvic X-rays that contain the hip. Diagnoses were extracted from radiology reports and curated by a practicing radiologist. Right hips were flipped across the vertical axis so that all hips appeared to be left hips.

To implement our model, we leveraged the code accompanying the original progressive GANs paper \citep{karras2017progressive}. We conducted several initial experiments (not shown) conditioning on simple features such as anatomic location and with smaller resolutions. Also not shown are additional experiments using a customization of the StackGAN architecture \citep{Han17stackgan2}, which yielded lower resolution images and worse conditioning. Due to resource constraints, the final GAN was only allowed to train for four days on two NVIDIA Titan GPUs, for a total of 8,800,000 images exposed to the discriminator.

\subsection{Experiments}

\subsubsection{Fracture Classification}
To assess whether our conditional GAN was properly conditioning on fractures, we conducted a series of experiments with classification models. To conduct these experiments, we split data our data into training (85\%) and test sets (15\%) by patient. We then trained our GAN exclusivley on the training set, and generated two additional "training sets" of the same number and class composition as the real training set by sampling from the conditional GAN with different seed values. For all classification models, Inception-v3 pretrained on ImageNet was fine-tuned after the last fully-connected layers replaced with a global average pooling, a fully-connected layer, and a sigmoid layer.

Our first test was a conditional variant of the boundary distortion test described in \citet{santurkar2017classification}. To execute this test, we first trained and compared classifiers on the real training set and as well as classifiers exclusively trained on each of the two GAN-sampled training sets. Each model was then evaluated on the test set of images held out from the classifier and GAN training.

To further test the characteristics of the GAN-sampled images, we trained classifiers using each of the GAN datasets as augmentation for the real data and for each other, as well as all datasets together. We additionally created one more baseline model by doubling the effective size of the real training set using traditional data augmentation.

\subsubsection{Nearest-Neighbors and t-SNE Analysis}
We next conducted experiments to test whether our model was merely memorizing images from the training set. To do so, we adapted the evaluation of \citet{chen2017photographic} and \citet{karras2017progressive}, comparing sampled images to near-neighbors in the training set. To find the nearest-neighbors of generated images, we trained VGG16 on the real images in the training and computing activation embeddings at each intermediate convolution layers. We then identified the nearest neighbor in the training set in L1 distance for each such layer. Finally, we utilized the same VGG model to generate t-SNE maps of 256 randomly sampled images from each class in each training set \citep{hinton2003stochastic}.

\begin{figure}
\begin{center}
        \includegraphics[scale=0.8]{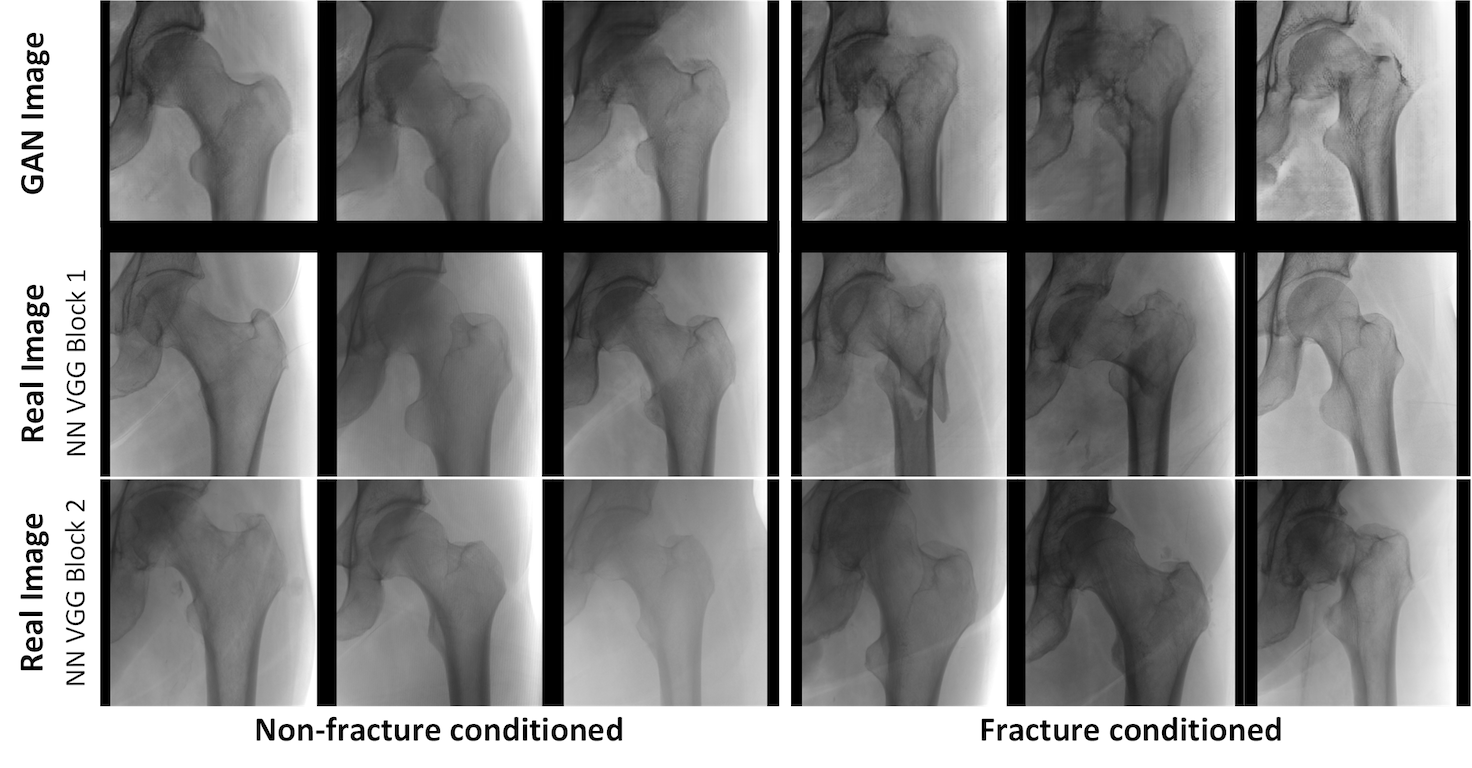}
        \caption{\label{fig:NNimages_first2rows} Randomly sampled images generated by the GAN (3 from each class using a pre-specified random seed), along with nearest-neighbors in the training set in feature space of a VGG16 model trained on radiographs.}
\end{center}
\end{figure}

\begin{table}[!htbp]
\centering
\scalebox{1.0}{
\begin{tabular}{c|c|c}
    \toprule
    Training data & AUC & AP \\
    \midrule
    \multicolumn{1}{r|}{Real} & 0.992 & 0.990 \\
    \multicolumn{1}{r|}{GAN-sample1} & 0.951 & 0.923 \\
    \multicolumn{1}{r|}{GAN-sample2} & 0.956 & 0.935 \\
    \multicolumn{1}{r|}{GAN-sample1 + GAN-sample2} & 0.945 & 0.924 \\
    \multicolumn{1}{r|}{Real + Augment} & 0.991 & 0.988 \\
    \multicolumn{1}{r|}{Real + GAN-sample1} & 0.991 & 0.988 \\
    \multicolumn{1}{r|}{Real + GAN-sample2} & 0.989 & 0.983 \\
    \multicolumn{1}{r|}{Real + GAN-sample1 + GAN-sample2} & 0.989 & 0.983 \\
    \bottomrule
\end{tabular}}
\caption{\label{table:results} Performance of fracture classification models trained with different sets of training data. All Area under the receiver operating curve (AUC) and average precision (AP) scores are reported on a held-out test set of real fracture images that were absent from the training set for the classifiers and GAN.}
\end{table}

\section{Results}

Results of our classification analysis is presented in Table \ref{table:results}. The highest test-set accuracy was achieved by training on real images (AUC 0.992), but all models achieved high performance, with models trained on GAN-sampled images alone achieving AUC $>0.95$.

Figure \ref{fig:NNimages_first2rows} depicts six randomly sampled (one time, with pre-specified seed) images from the GAN, along with top-1 near-neighbors in the feature space from the first two convolutional layers of a VGG network trained on fracture images. Top-1 neighbors from all other layers are in the appendix in Figure \ref{fig:NNimages}. No obvious copies are visible in any case. Figures 2-4 in the appendix depict t-SNE plots from 256 images sampled from each class from the real and first GAN-sampled training set. No obvious mode collapse is present, and a range of different types of fractures and annotation markers were learned by the GAN, though the samples appear less diverse than the true images.

\section{Discussion}

Visual inspection and classification results both indicate that our conditional GAN has learned key visual features that distinguish fracture from non-fracture images. Though of slightly lower quality than real images in appearance and performance, the objective success of classifiers trained solely on GAN samples is still striking and strongly implies that our GAN has learned clinical-relevant features sufficient to aid in machine learning. Based on these results, we are currently designing a simple trial to test the extent to which GAN-sampled images serve as an effective alternative to real images in \textit{human} learning. Of note, while simulating the true data distribution is the goal of GAN training, in the setting of medical education, whether or not sampled images are indistinguishable from or as diverse as real images may or may not prove to fully correlated with whether they are helpful to students learning to internalize the key distinction between fractures from non-fractures.

We are particularly intrigued by the data augmentation results. Prior studies in the medical domain have reported improved classification performance using GANs as a means of data augmentation, however none of these in the setting of nearly perfect performance on real images alone \citep{santurkar2017classification, li2018limitations, antoniou2017data}. These findings seem to suggest that GAN images may serve as effective augmentation in the low-data regime, but may be actively unhelpful in a high-data regime. We presently intend to conduct experiments simulating the relative impact of GAN augmentation when progressively limiting access to real images. 

\section{Acknowledgements}

SGF was supported by training grants T32GM007753 and T15LM007092; the content is solely the responsibility of the authors and does not necessarily represent the official views of the National Institute of General Medical Sciences or the National Institutes of Health.

\bibliographystyle{ACM-Reference-Format}
\bibliography{fracgan}

\newpage

\section*{Appendix}

\begin{figure}[!htb]
  \centering
    \includegraphics[width=\textwidth]{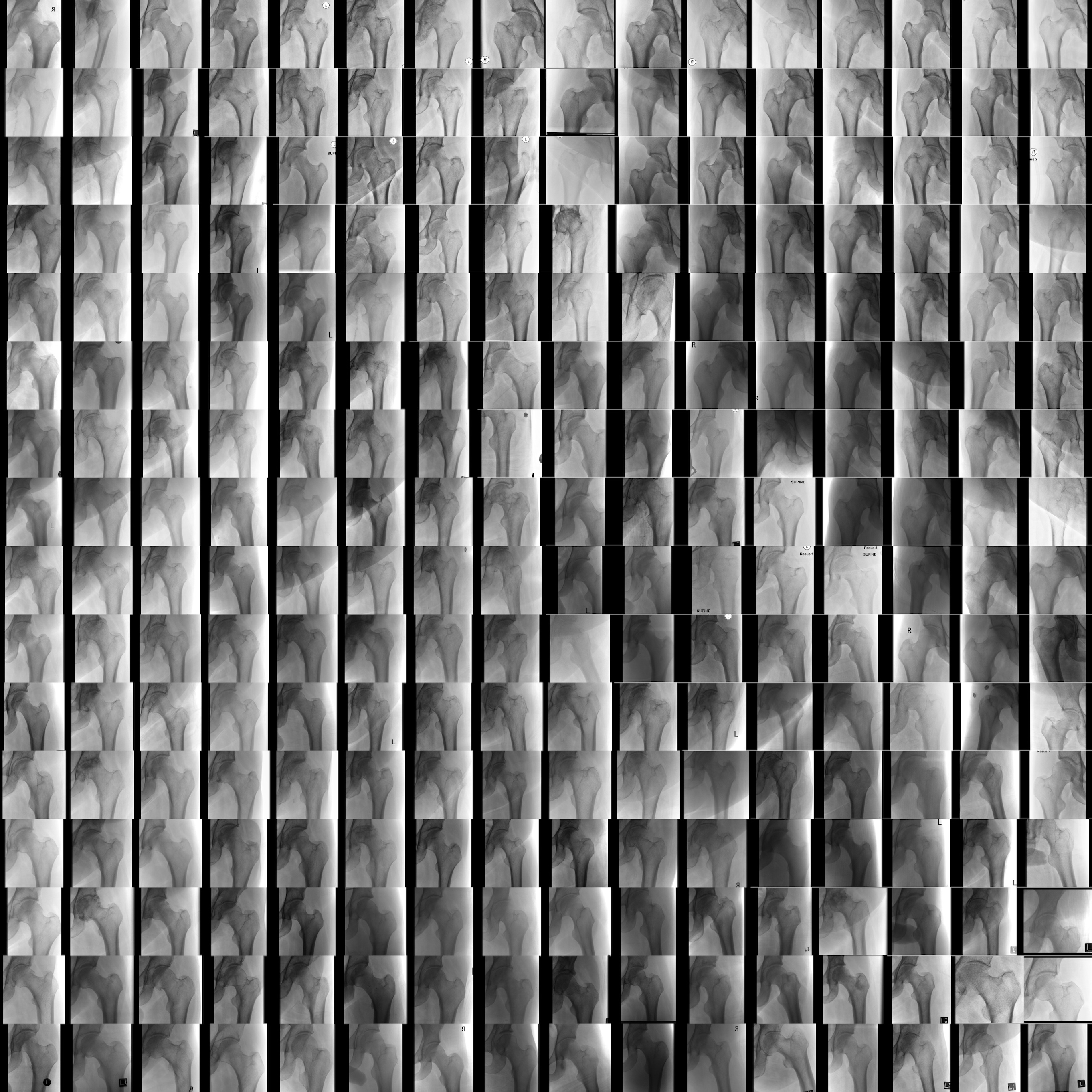}
    \caption{\label{fig:real_nonfrac} t-SNE plot of additional examples of \textbf{real, non-fractured} images.}
\end{figure}

\begin{figure}[!htb]
  \centering
    \includegraphics[width=\textwidth]{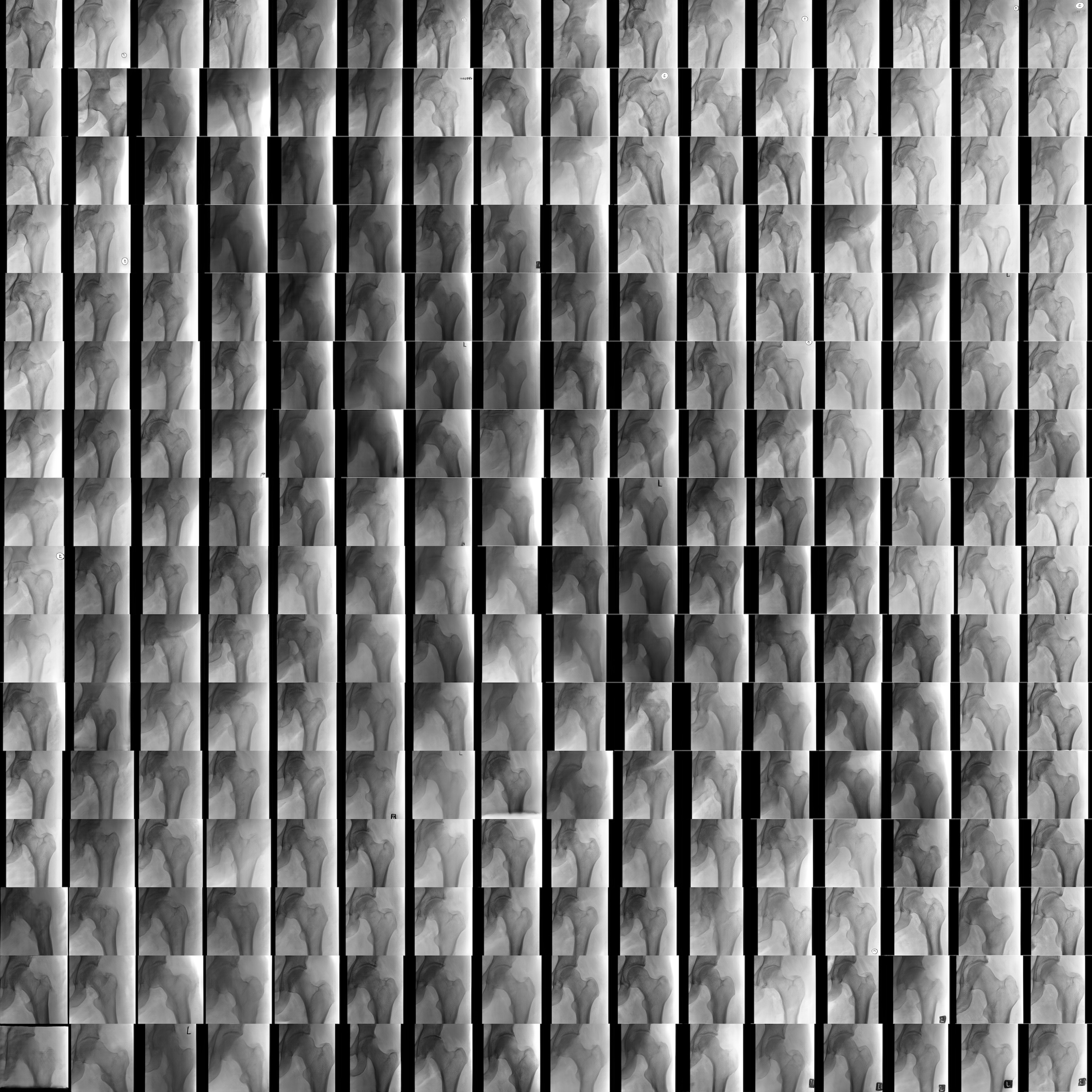}
    \caption{\label{fig:appen_gan_nonfr} t-SNE plot of additional examples of \textbf{GAN-generated, non-fractured} images.}
\end{figure}

\begin{figure}[!htb]
  \centering
    \includegraphics[width=\textwidth]{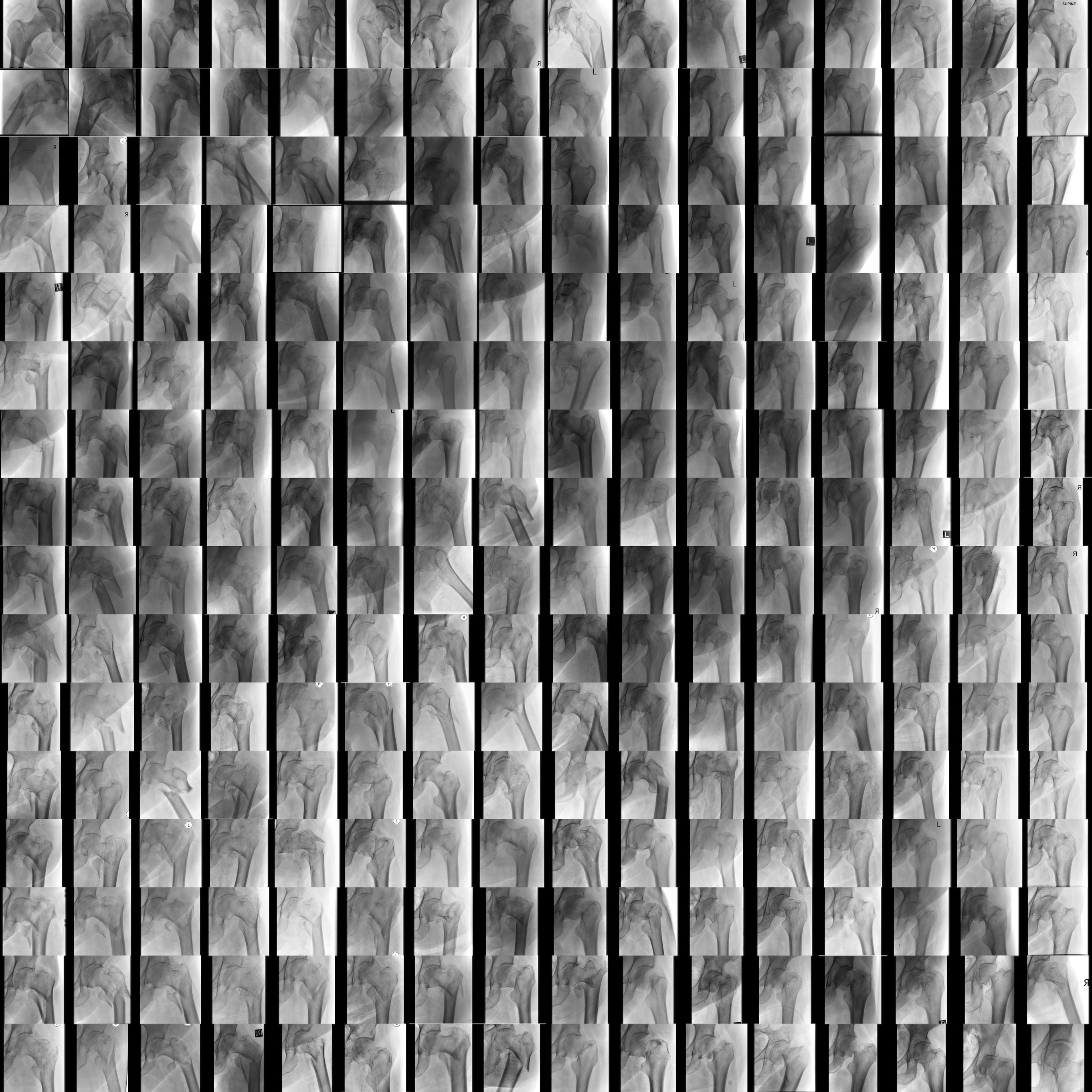}
    \caption{\label{fig:real_frac} t-SNE plot of additional examples of \textbf{real, fractured} images.}
\end{figure}

\begin{figure}[!htb]
  \centering
    \includegraphics[width=\textwidth]{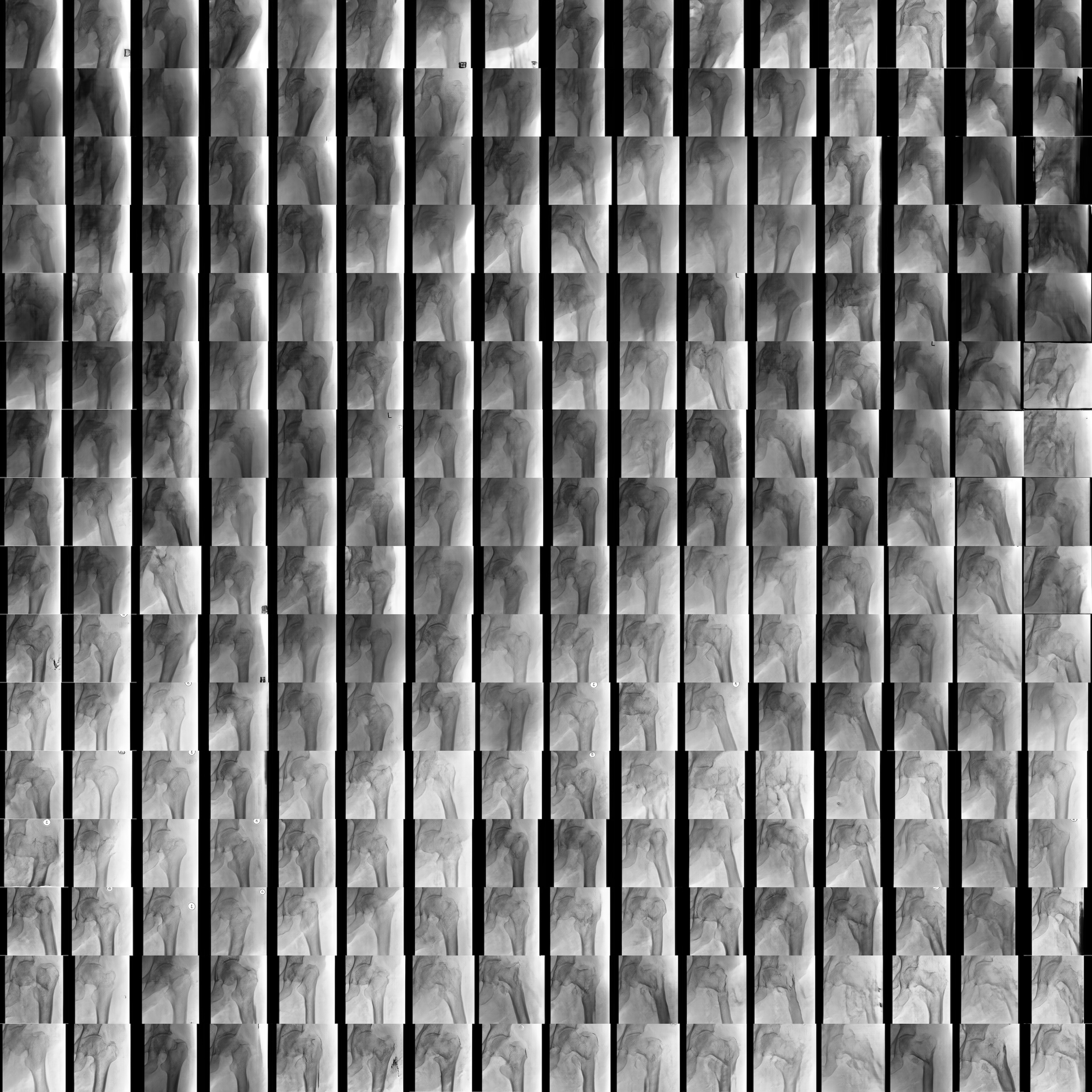}
    \caption{\label{fig:appen_gan_fr} t-SNE plot of additional examples of \textbf{GAN-generated, fractured} images.}
\end{figure}

\begin{figure}[!htb]
  \centering
    \includegraphics[width=\textwidth]{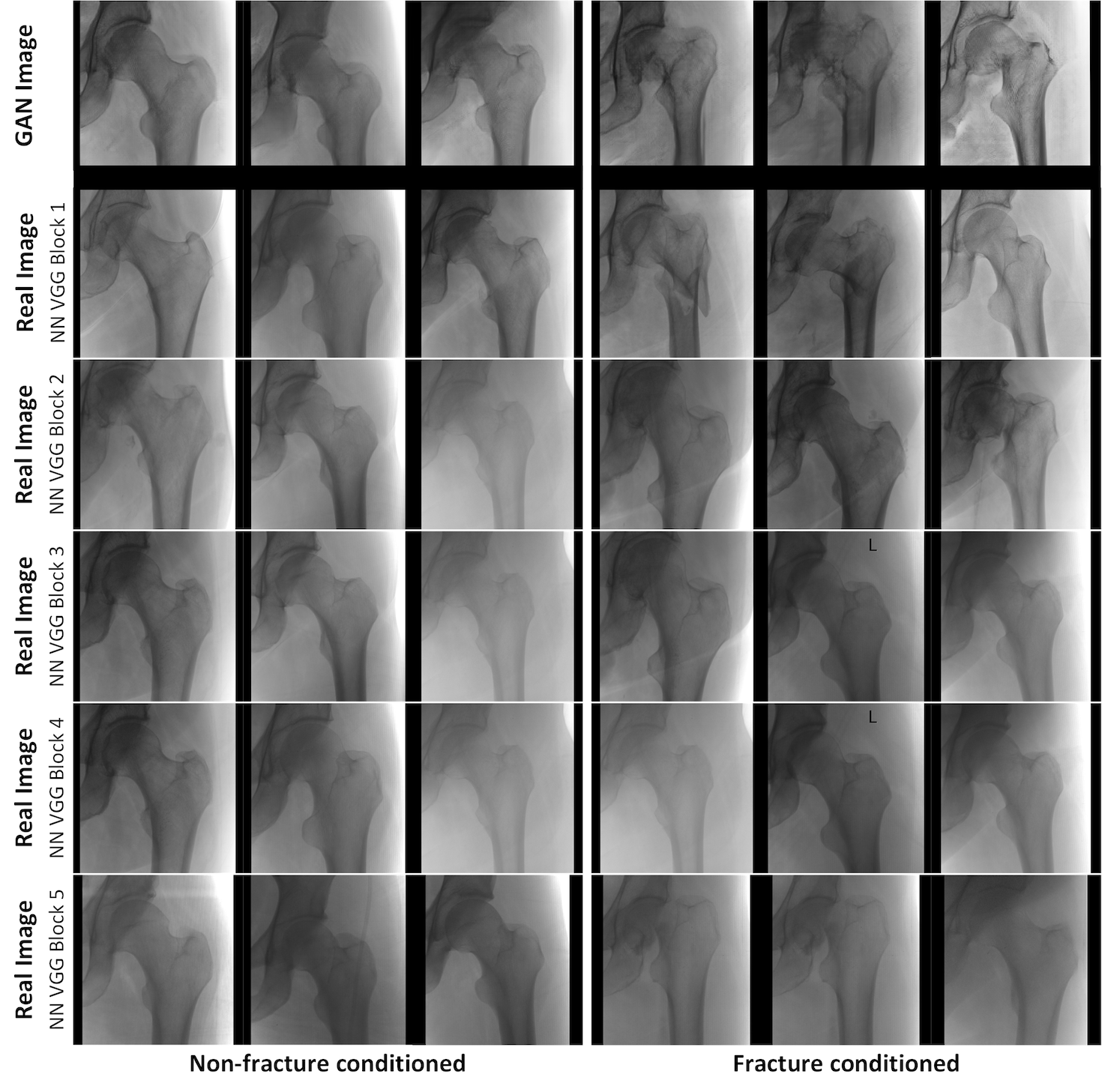}
    \caption{\label{fig:NNimages} t-SNE plot of additional examples of \textbf{real and GANq    } images.}
\end{figure}

\end{document}